\documentclass{article}
\usepackage{spconf,amsmath,graphicx}
\usepackage{balance}
\usepackage{xcolor}
\usepackage{xurl}

\title{Innovative Methods for Non-Destructive Inspection\\of Handwritten Documents}
\name{Eleonora Breci$^{\star}$, Luca Guarnera$^{\star}$, Sebastiano Battiato$^{\star}$}
\address{$^{\star}$ Department of Mathematics and Computer Science, University of Catania, Italy \\
\emph{brecieleonora@gmail.com, \{luca.guarnera, sebastiano.battiato\}@unict.it}}
\begin{document}
%
\maketitle
\begin{abstract}
Handwritten document analysis is an area of forensic science, with the goal of establishing authorship of documents through examination of inherent characteristics. Law enforcement agencies use standard protocols based on manual processing of handwritten documents. This method is time-consuming, is often subjective in its evaluation, and is not replicable. To overcome these limitations, in this paper we present a framework capable of extracting and analyzing intrinsic measures of manuscript documents related to text line heights, space between words, and character sizes using image processing and deep learning techniques. 
The final feature vector for each document involved consists of the mean ($\eta$) and standard deviation ($\sigma$) for every type of measure collected. By quantifying the Euclidean distance between the feature vectors of the documents to be compared, authorship can be discerned. 
Our study pioneered the comparison between traditionally handwritten documents and those produced with digital tools (e.g., tablets). Experimental results demonstrate the ability of our method to objectively determine authorship in different writing media, outperforming the state of the art. 
\end{abstract}

\section{Introduction}
Forensic handwriting examination is a branch of Forensic Science that aims to examine handwritten documents in order to define the author of the manuscript. These analysis involves comparing two or more (digitized) documents through a comprehensive comparison of intrinsic features. 
The need to create sophisticated tools capable of extracting and comparing significant features has led to the development of cutting-edge software with almost entirely automated processes, improving the forensic examination of handwriting and achieving increasingly objective evaluations. This is made possible by algorithmic solutions based on purely mathematical concepts.


\begin{figure*}[t!]
    \centering     \includegraphics[width=\linewidth]{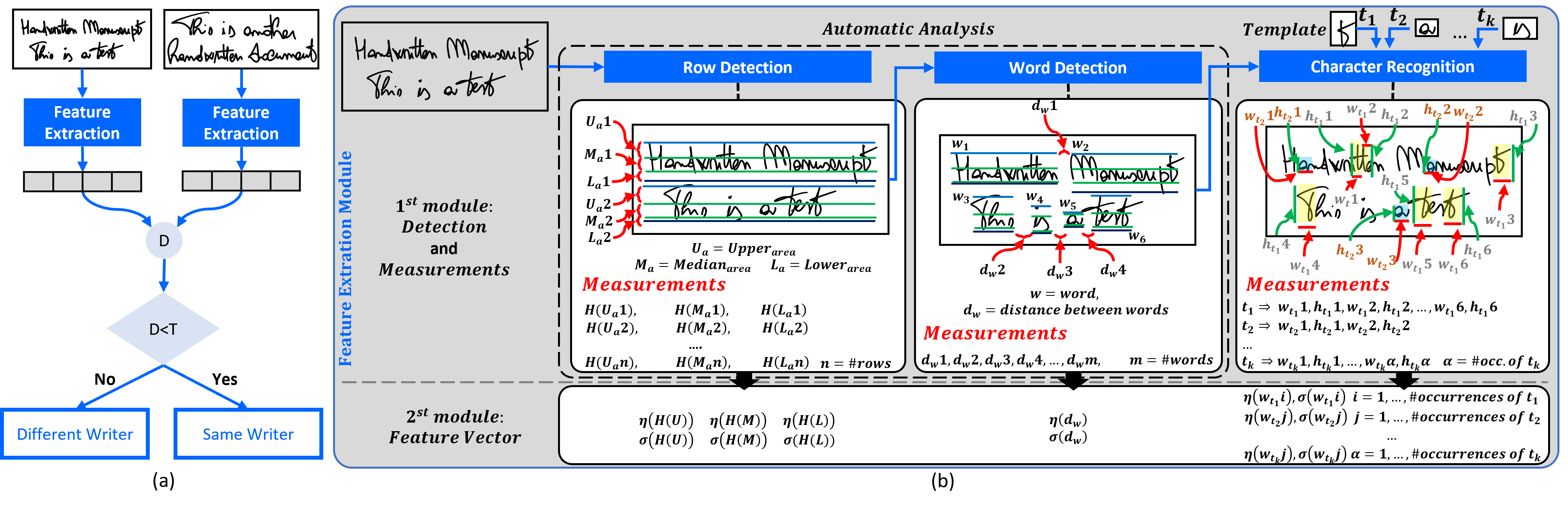}\vspace{-0.3cm}
    \caption{Proposed approach. (a) Comparison between documents. (b) Feature extraction module: $1^{st}$ module): lines of text and words are automatically detected. Then, one or more characters (templates) chosen by the expert can be searched within the document. $2^{st}$ module): the feature vector is defined as the means ($\eta$) and standard deviations ($\sigma$) of all collected measures.}
    \vspace{-0.3cm}
    \label{fig:featureextactionscheme}
\end{figure*}

The goal of this work was to develop a new, more objective, repeatable and automatic scientific method for analyzing manuscript documents to define the authorship. Specifically, given two or more manuscript documents under analysis, the proposed approach is able to detect all measures related to text lines and words, such as heights of upper, middle and lower areas, space between words and so on. A Deep Learning approach was developed to recognize a specific character, called template, chosen by the forensic expert in the document. Then, the height and width measurements of all detected occurrences are calculated. For each document, a final feature vector is then created consisting of \textit{Means} ($\eta$) and \textit{Standard Deviations} ($\sigma$) calculated for each type of extracted measures. 
Finally, to define authorship, the Euclidean distance between two feature vectors obtained from two different manuscripts is calculated. 
For the first time in this field, the comparison of documents written with both the classic ``pen and paper" approach and written with digital devices (e.g., tablet) was performed and tested. The obtained results in the experimental phase show that the proposed framework is able to infer the identity of the writer, not only by comparing only handwritten documents and later digitized to perform forensic analysis, but also by comparing documents written directly on digital devices. Moreover, even by comparing a document written on paper with one written on a tablet, we are able to define whether the writer is the same. This approach innovates by automating analysis and delivering a more objective final assessment. 



This paper is organized as follows: Section~\ref{sec:sota} presents some state-of-the-art approaches in this area. Section~\ref{sec:method} describes the proposed algorithms. Experimental results and comparison are presented in Section~\ref{sec:exp}. Finally, Section~\ref{sec:conclusion} concludes the paper with hints for future work.

\vspace{-0.3cm}
\section{Related Works}
\label{sec:sota}

Several studies~\cite{deviterne2020interpol,morris2020forensic,koppenhaver2007forensic} have analyzed the problem of writing identification with the aim of defining the most discriminating features to assess the identity of the writer. Morris~\cite{morris2020forensic} emphasized the importance of evaluating dimensional parameters by comparing absolute and relative quantities in order to analyze speed, slant, and style and detect possible attempts at forgery. According to Koppenhaver~\cite{koppenhaver2007forensic}, the analysis of character heights can provide discriminative information about an individual's handwriting variability. 
Guarnera et al.~\cite{guarnera2017graphj,guarnera2018forensic} showed that the relationship between certain intrinsic measures (e.g., distance between words and between letters, height and width of characters, etc.) can be used to identify the person who wrote a text.

Recent state-of-the-art approaches have addressed the task of identifying authors using Machine Learning and Deep Learning techniques on different datasets. Crawford et al.~\cite{crawford2023statistical} presented a statistical approach to model various handwriting styles, aiming to calculate the likelihood of a questioned document's authorship from a known set of writers. 
Semma et al.~\cite{semma2021writer} extracted key points from handwriting and using Convolutional Neural Networks (CNNs) for classification. FAST key points and Harris corner detectors identify points of interest, and a CNN is trained on patches around these points. Bennour et al.~\cite{bennour2019handwriting} explored writer characterization in handwriting recognition using an implicit shape codebook technique. Key points in handwriting are identified and used to create a codebook, leading to promising results in various experimental scenarios. Kumar et al.~\cite{kumar2020segmentation} introduced a writer identification model (SEG-WI) using a CNN and weakly supervised region selection. It achieves segmentation-free writer identification on various datasets, outperforming state-of-the-art methods. Lai et al.~\cite{lai2020encoding} proposed novel techniques using pathlet and unidirectional SIFT features for fine-grained handwriting description. He et al.~\cite{he2020fragnet} presented FragNet, a deep neural network with two pathways to extract powerful features for writer identification from word and page images. Bahram~\cite{bahram2022texture} used co-occurrence features extracted from preprocessed regions of interest and contour texture.

\vspace{-0.2cm}
\section{Proposed Approach}
\label{sec:method}
Four different algorithms have been implemented for: (a) Text line detection; (b) Word detection; (c) Character recognition; and (d) Manuscript comparison.  The first two methods (a,b) are performed automatically after the manuscript to be analyzed is given as input to the framework. Automatic character recognition occurs when the forensic operator selects the template (i.e., a character in the manuscript itself) to be searched. Finally, once the feature vectors of two handwritten documents have been extracted, they can be compared to define authorship. 

\vspace{-0.4cm}
\subsection{Text Line Detection} 
\begin{figure}[t!]
    \centering     \includegraphics[width=\linewidth]{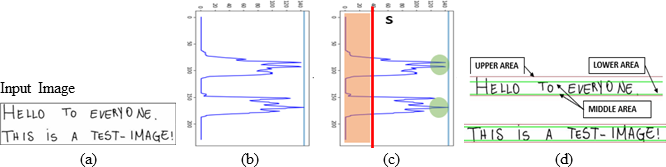}\vspace{-0.3cm}
    \caption{(a) Text line detection algorithm of the input binarized image $I_B$. (b) Computation of histogram $H_{row}$ from $I_B$ . (c) Text line detection (each observed peak). (d) Search for the upper, middle and lower areas.} 
    \vspace{-0.3cm}
    \label{fig:hist}
\end{figure}

The text line detection algorithm initially involves the binarization operation ($I_B$) of the input image ($I$). An histogram $H_{row}$ containing the number of occurrences of black pixels is computed by $I_B$ and analyzed to detect text lines (Figure~\ref{fig:hist} (b)). To best solve the proposed task, the noise present in $H_{row}$ is removed. Specifically, a threshold $S$ is computed as the average of the values in $H_{row}$, and any value in $H_{row}$ less than $S$ is set to 0 and consequently excluded in the search for subsequent text lines (Figure~\ref{fig:hist} (c) red area). Then, to detect a line of text (Figure~\ref{fig:hist}(c)) the maximum value in $H_{row}$ is detected and the index position $ind_{max}$ is saved. For each line of text detected, the upper, middle and lower area is identified, as shown in Figure~\ref{fig:hist} (d). For this purpose, a $val = max/4$ is set, as described in~\cite{guarnera2017graphj,guarnera2018forensic}. The variable $x$ is set equal to $ind_{max}$ and decremented by 1 until $H_{row}[x] < val$. At condition $False$ the beginning of the middle area is defined. Vice versa, to search for the end index of the middle area, $x$ is set equal to $ind_{max}$ and incremented by 1 until $H_{row}[x] < val$.  Under the condition $False$ the end of the central area is defined.
Iteratively, all lines of text in the manuscript will be detected.
Then, for each line of text detected, the top and bottom areas are searched.
Specifically, given a generic identified line of text, we set a variable $x$ with the index of the beginning of the middle area. If $H_{row}[x]==0$ then $x$ will be the index of the beginning of the upper area, otherwise $x$ is decremented by 1 until the condition ($H_{row}[x]==0$) is satisfied. In the case that this condition is never satisfied and $x$ points to the end of the middle area of the previous line of text in the manuscript or if $x == 0$ (meaning in this case that we are in the first row of pixels of $I_B$), then the beginning of the upper area will be defined as the minimum value in the range between $H_{row}[x]$ and $H_{row}$ of the beginning of the middle area of the line of text under analysis. The reverse process is applied to find the end of the lower area.
The height between the beginning and end of each detected area is stored and used to create the final feature vector.
These represent discriminant measures capable of defining the author of the manuscript.

\vspace{-0.3cm}
\subsection{Word Detection}

To identify all words in the manuscript, an histogram of columns $H_c$, defined as the sum of black pixels per column, is created for each detected (and binarized) line of text. We define space between two words as the number of occurrences of consecutive zeros in $H_c$ greater than a predefined threshold value $So$. Values of $H_c$ not included in such sequences of zeros define the words in the text row under analysis. 
In cases where some words are not detected correctly, the threshold value $So$ can be changed manually to correct the obtained result. The space between the detected words will be stored and used to create the final feature vector.

\vspace{-0.3cm}
\subsection{Character Recognition}

\begin{figure}[t!]
    \centering     \includegraphics[width=\linewidth]{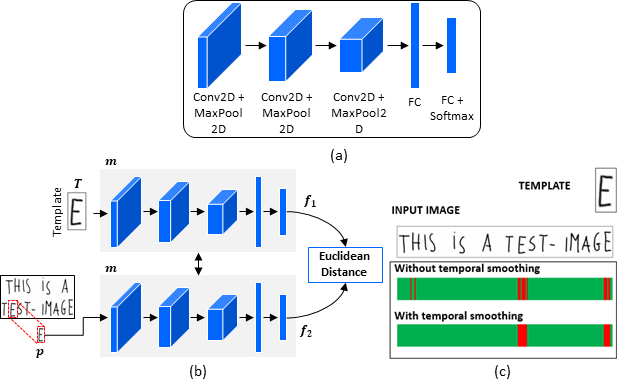}\vspace{-0.3cm}
    \caption{(a) Architecture of the proposed deep neural network. (b) $T$ and $p$ are analyzed by $m$, and the Euclidean distance between the two respective feature vectors is calculated. (c) A temporal smoothing operation is applied to remove characters detected incorrectly during the search.} \vspace{-0.3cm}
\label{fig:charcacter}
\end{figure}

A deep learning approach was considered for automatic character recognition. Initially, a deep neural network (implemented using the Keras~\footnote{Keras library: https://keras.io/}) consisting of 3 Conv2D and 3 MaxPool2D layers, 1 fully connected layer with output size $128$ and a last fully connected layer with output size $36$ followed by a SoftMax (Figure~\ref{fig:charcacter} (a)) was implemented.
The architecture was trained for $30$ epochs using the Adam optimizer with a learning rate of $0.0001$. The \textit{A Z Handwritten Data} ~\cite{A-Z-Handwritten} 
dataset was considered to train and test the model. 
The network was trained to classify $36$ different characters and numbers. The dataset was divided into training ($60\%$), validation ($20\%$) and test ($20\%$) sets.

The best model ($m$) achieved in the test phase a classification accuracy equal to $99,08\%$. 
Once this phase was completed, $m$ was configured in Siamese modality to solve the proposed approach (Figure~\ref{fig:charcacter} (b)). The forensic expert selects the character (of size $(h,w)$) to be searched in the manuscript, called template ($T$). A sliding window moves (one pixel at a time) along the area of each identified line of text and a patch $p$ having the same size as $T$ is extracted. $T$ and $p$ will be analyzed by $m$ obtaining the feature vectors $f_1$ and $f_2$, respectively. The Euclidean distance ($d = D(f_1,f_2)$ (Eq.~\ref{eq:eucl})) will be calculated to define the similarity between $T$ and $p$. If $d$ is less than an empirical threshold ($T_c = 0.1$), then $T$ and $p$ represent the same information.

Once all patches are classified, a temporal smoothing operation will be applied to proceed with the elimination of false positives. When at least 5 consecutive comparisons satisfy the equality between $T$ and $p$, the character in $T$ can be considered found (Figure~\ref{fig:charcacter} (c)).
The height and width of all found occurrences are stored to define the final feature vector.

\begin{equation}
    D(x,y) = \sqrt{\sum_{i=1}^{n}(x_i-y_i)^2}
    \label{eq:eucl}
\end{equation}

\vspace{-0.5cm}
\subsection{Feature Vector Creation}

For each document, a final feature vector is created consisting of \textit{Means} ($\eta$) and \textit{Standard Deviations} ($\sigma$) calculated for each type of extracted measure: \textbf{(i)} height of upper, middle, and lower areas of text lines; \textbf{(ii)} space between words; \textbf{(iii)} height and width of each character detected for each specific template (Figure~\ref{fig:featureextactionscheme} ($2^{st}$ module)).

\section{Experimental Results and Comparison}
\label{sec:exp}

\begin{figure}[t!]
    \centering     \includegraphics[width=\linewidth]{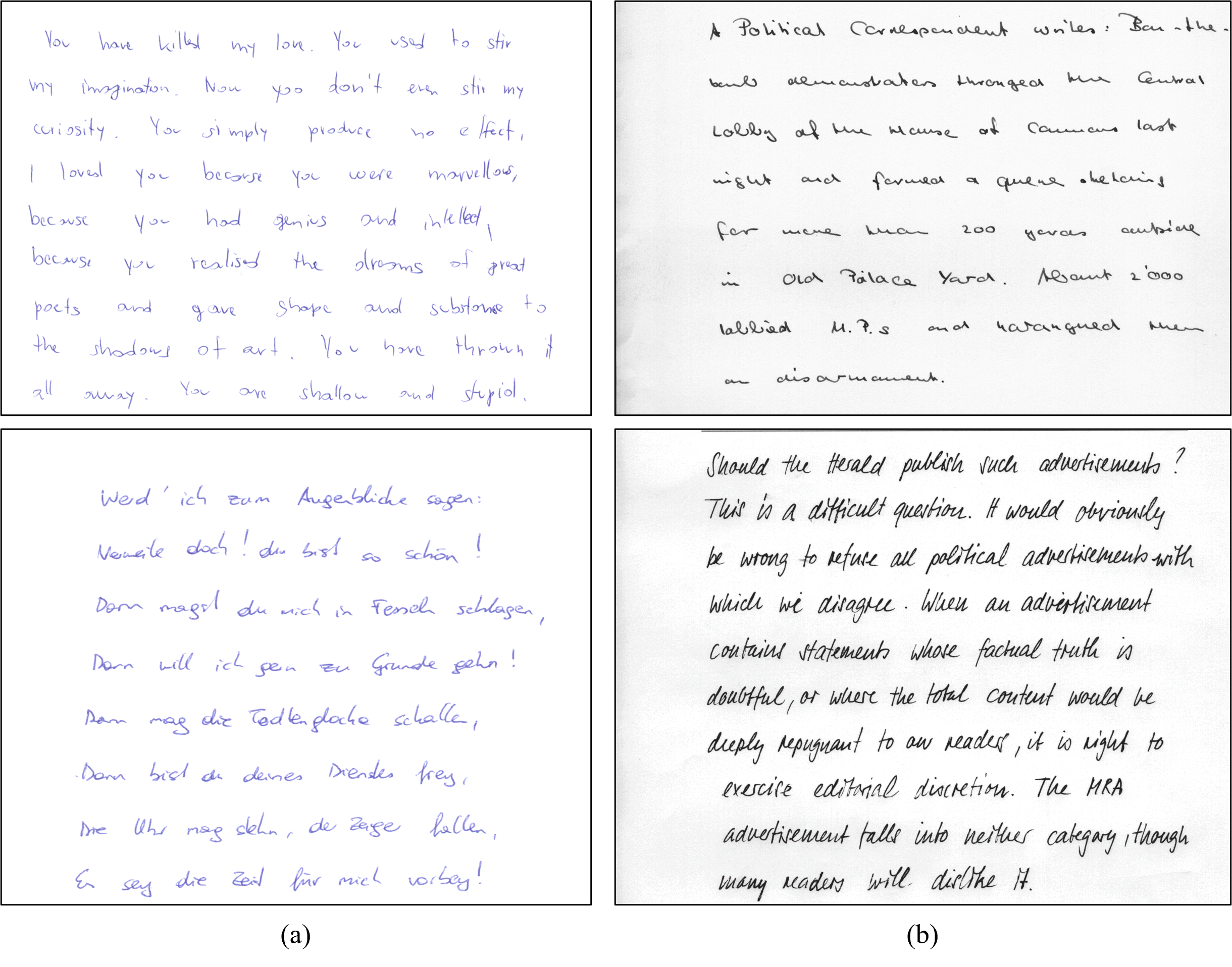}\vspace{-0.4cm}
  \caption{Examples of (a) CVL \cite{kleber2013cvl} and (b) CSAFE \cite{crawford2020database} digitized texts.}
\label{fig:cvl-iam-dataset}
\end{figure}
\begin{figure}[t!]
    \centering     \includegraphics[width=\linewidth]{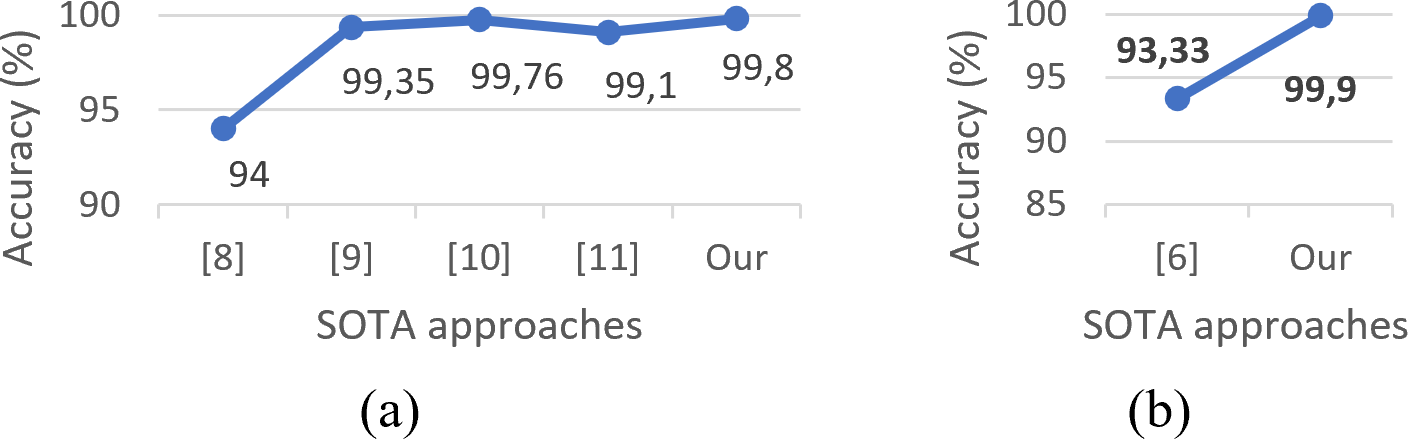}\vspace{-0.4cm}
    \caption{Comparison with state-of-the-art (SOTA) approaches, by using CVL (a) and CVL + CSAFE datasets (b).}
\label{fig:comparison}
\end{figure}
The proposed method was tested with the following datasets most used by the scientific community:
\textbf{(i)} Computer Vision Lab (CVL) Database \cite{kleber2013cvl}, includes handwritten text from 311 unique writers, comprising both English and German samples. We selected a total of 1600 handwritten documents. \textbf{(ii)} Center for Statistics and Applications in Forensic Evidence (CSAFE) Handwriting Database \cite{crawford2020database}, that consist of a total of 2430 handwriting sample images collected from surveys of 90 writers.
Figure \ref{fig:cvl-iam-dataset} shows some examples of these datasets. As reported in Figure \ref{fig:featureextactionscheme} (a), two documents were compared at a time (considering all possible combinations) in order to assert the identity of the writer. 
In experimental tests with the CVL dataset, we obtained a classification accuracy of $99.8\%$. Figure \ref{fig:comparison} (a) shows the results obtained compared with some recent state-of-the-art (SOTA) methods~\cite{bennour2019handwriting,kumar2020segmentation,lai2020encoding,he2020fragnet}. 
In this context, SOTA approaches are able to identify the writer well. Our approach is able to achieve the best results. 
Considering the CSAFE dataset, we obtained a classification accuracy of $100\%$.  \cite{crawford2023statistical} analyzed \cite{crawford2020database,kleber2013cvl} together (as one dataset). Considering this configuration, our approach achieved a classification accuracy of $99.9\%$ exceeding \cite{crawford2023statistical} ($93.33\%$) (Figure~\ref{fig:comparison}~(b)). Compared with the various works in the literature just cited, the proposed method appears to be better, repeatable, more objective (in terms of how the feature vector was defined), and can be used by law enforcement agencies in real-world scenarios.

\begin{figure}[t!]
    \centering     \includegraphics[width=\linewidth]{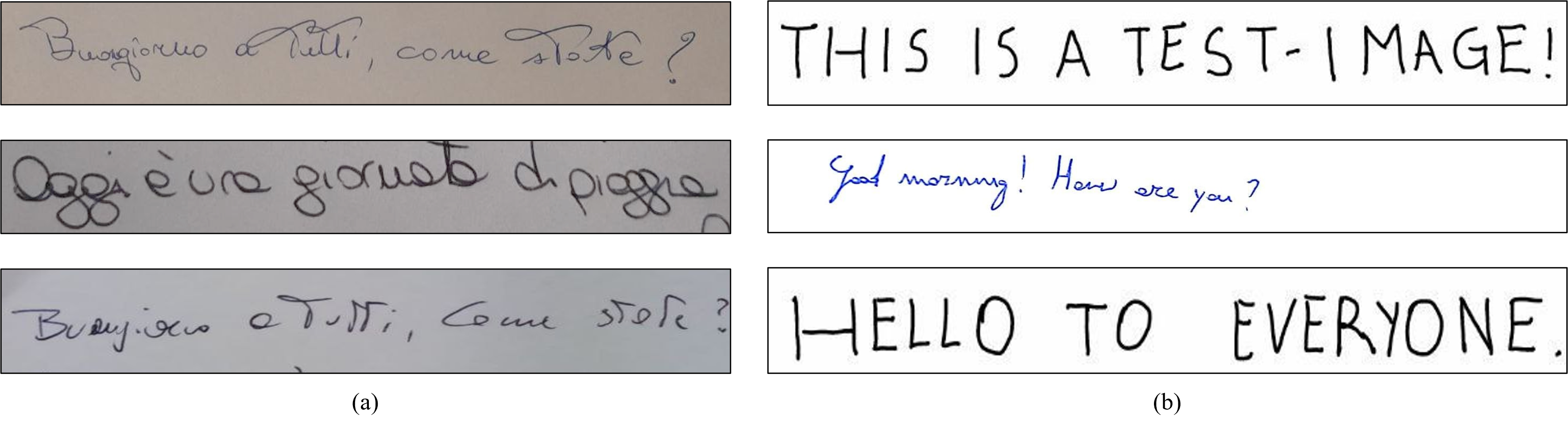}\vspace{-0.3cm}
    \caption{Sample images of the new dataset of handwritten manuscripts using the ``pen and paper" approach (a) and digital devices (b).}\vspace{-0.3cm}
\label{fig:newsdataset}
\end{figure}

Finally, to evaluate the discriminative power of the constructed feature vector, we created a new challenging dataset~\cite{breci2024anovel} consisting of 362 manuscript documents, all written by 124 different authors. These manuscripts (Figure \ref{fig:newsdataset}) included sentences with different languages and types of writing, including bold or italics. In addition, for the first time in this field, documents both written with the classic ``pen and paper" approach (and later digitized) and written with digital devices (e.g., tablet) were collected. 
All documents were compared with each other, resulting in a classification accuracy 
of~$96\%$. 

\vspace{-0.2cm}
\section{Conclusion and Future Works}
\label{sec:conclusion}

In this paper, a new framework was developed for the analysis and comparison of handwritten documents capable of extracting discriminative measurements through the automatic analysis of text lines, words, and characters. 
For each document under analysis, a feature vector was defined consisting of mean ($\eta$) and standard deviations ($\sigma$) calculated for each type of extracted measurement. The experimental results demonstrate that the proposed method allows the extraction of quantitative information able to infer the identity of the writer.
For the first time in this field, a comparison was made between documents written with both the classic ``pen and paper" approach (and later digitized) and those written with digital devices (e.g., tablet).  The proposed framework provides a more objective and automatic method, improving the Forensic examination of handwritten manuscripts.
Future work will be oriented toward a more detailed description and extension of the involved 
dataset 
containing many more examples of traditionally handwritten documents and those produced with digital tools.
In addition, other deep learning algorithms will be implemented in order to improve existing solutions and obtain even more accurate results. The code and datasets are available at \url{https://iplab.dmi.unict.it/mfs/forensic-handwriting-analysis/innovative-methods-2023/}. 

\section{Acknowledgments} 
This research is supported by Azione IV.4 - ``Dottorati e contratti di ricerca su  tematiche dell’innovazione" del nuovo Asse IV del PON Ricerca e Innovazione 2014-2020 “Istruzione e ricerca  per il recupero - REACT-EU”- CUP: E65F21002580005.

\balance{
\bibliographystyle{IEEE}
\bibliography{references}
}

\end{document}